\documentclass[sigconf,screen]{acmart}

\usepackage{amsmath}
\usepackage{graphicx}
\usepackage{multirow}
\usepackage{array}
\usepackage{indentfirst}
\usepackage{svg}
\usepackage{float}

\usepackage{algorithm}
\usepackage{algpseudocodex}
\usepackage{setspace}

\AtBeginDocument{%
  \providecommand\BibTeX{{%
    \normalfont B\kern-0.5em{\scshape i\kern-0.25em b}\kern-0.8em\TeX}}}

\setcopyright{acmlicensed}
\copyrightyear{2024}
\acmYear{2024}
\acmDOI{XXXXXXX.XXXXXXX}

\acmConference[MMSys '24]{Multimedia Systems}{April 15--18,
  2024}{Bari, Italy}
\acmSubmissionID{59}

\begin{document}
\acmYear{2024}\copyrightyear{2024}
\acmConference[MMSys '24]{ACM Multimedia Systems Conference 2024}{April 15--18, 2024}{Bari, Italy}
\acmBooktitle{ACM Multimedia Systems Conference 2024 (MMSys '24), April 15--18, 2024, Bari, Italy}
\acmDOI{10.1145/3625468.3647620}
\acmISBN{979-8-4007-0412-3/24/04}
\title{Finding Waldo: Towards Efficient Exploration of NeRF Scene Spaces}


\author{Evangelos Skartados} \authornote{Equal contribution.}
\email{eskartad@iti.gr}
\affiliation{%
  \institution{CERTH ITI}
  \city{Thessaloniki}
  \country{Greece}
}

\author{Mehmet Kerim Yucel} \authornotemark[1]
\email{mehmet.yucel@samsung.com}
\affiliation{%
  \institution{Samsung Research UK}
  \streetaddress{Communications House, South St}
  \city{Staines}
  \state{Surrey}
  \country{United Kingdom}
  \postcode{TW18 4QE}
}

\author{Bruno Manganelli}
\email{b.manganelli@samsung.com}
\affiliation{%
  \institution{Samsung Research UK}
  \streetaddress{Communications House, South St}
  \city{Staines}
  \state{Surrey}
  \country{United Kingdom}
  \postcode{TW18 4QE}
}

\author{Anastasios Drosou}
\email{drosou@iti.gr}
\affiliation{%
  \institution{CERTH ITI}
  \city{Thessaloniki}
  \country{Greece}
}

\author{Albert Sa\`a-Garriga}
\email{a.garriga@samsung.com}
\affiliation{%
  \institution{Samsung Research UK}
  \streetaddress{Communications House, South St}
  \city{Staines}
  \state{Surrey}
  \country{United Kingdom}
  \postcode{TW18 4QE}
}

\renewcommand{\shortauthors}{Skartados and Yucel, et al.}

\begin{abstract}
 Neural Radiance Fields (NeRF) have quickly become the primary approach for 3D reconstruction and novel view synthesis in recent years due to their remarkable performance. Despite the huge interest in NeRF methods, a practical use case of NeRFs has largely been ignored; the exploration of the scene space modelled by a NeRF. In this paper, for the first time in the literature, we propose and formally define the scene exploration framework as the efficient discovery of NeRF model inputs (i.e. coordinates and viewing angles), using which one can render novel views that adhere to user-selected criteria. To remedy the lack of approaches addressing scene exploration, we first propose two baseline methods called Guided-Random Search (GRS) and Pose Interpolation-based Search (PIBS). We then cast scene exploration as an optimization problem, and propose the criteria-agnostic Evolution-Guided Pose Search (EGPS) for efficient exploration. We test all three approaches with various criteria (\textit{e.g.} saliency maximization, image quality maximization, photo-composition quality improvement) and show that our EGPS performs more favourably than other baselines. We finally highlight key points and limitations, and outline directions for future research in scene exploration.
\end{abstract}



\begin{teaserfigure}
    \centering
  \includegraphics[width=0.92\textwidth]{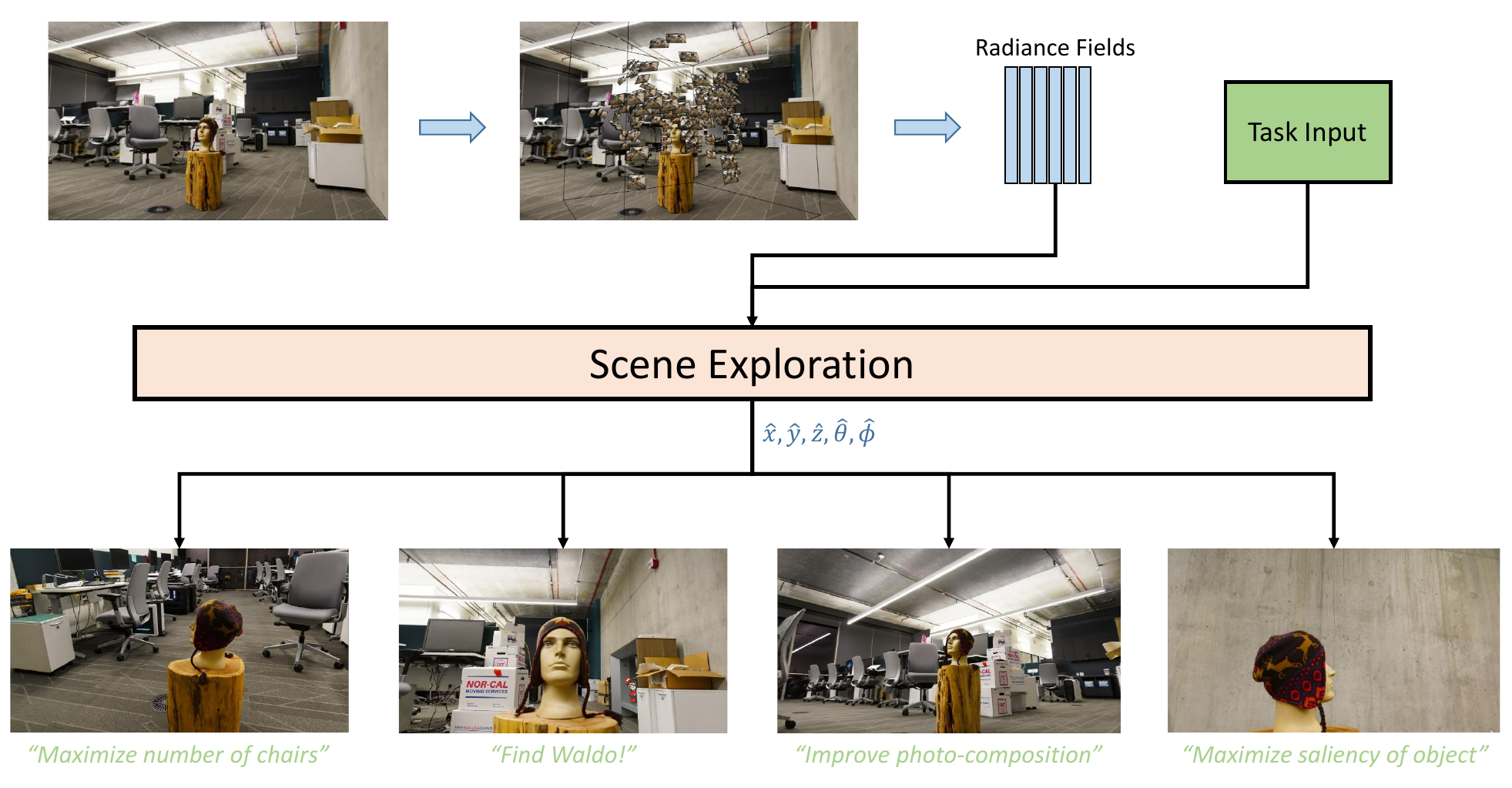}
  \vspace{-4mm}
  \caption{In this work, we formally define and introduce the scene exploration framework; given a scene and a NeRF model that encodes it, the scene exploration framework aims to find the camera poses from which one can render novel views that adhere to some user-provided criteria; \textit{e.g.} including an object, improving photo-composition, maximizing object saliency, etc. Efficiently exploring scenes in 3D can be imperative for content creation, multimedia production and VR/AR applications. Refer to Algorithm \ref{algo:baselines} for a detailed breakdown.}
  \label{fig:teaser}
\end{teaserfigure}


\maketitle

\section{Introduction} \label{introduction}

\noindent  Neural Radiance Fields (NeRF) have emerged as a strong alternative for compactly representing 3D scenes and novel view synthesis, and quickly became a popular tool for 3D reconstruction, thanks to their superior scene modelling ability and memory-efficiency. NeRF methods have seen tremendous interest from the community, with many follow-up works introducing improvements in various aspects \cite{barron2021mip,deng2022depth,muller2022instant,reiser2021kilonerf,yang2023freenerf,jain2021putting}. Furthermore, NeRF has swiftly found use in numerous applications,  such as robotics \cite{adamkiewicz2022vision}, augmented reality \cite{li2023instant} and computational photography \cite{mildenhall2021nerf}, with many more potential applications waiting for NeRF adoption.

Many visual tasks that are traditionally performed on 2D images are adapted to work with NeRF representations as well, such as segmentation and detection \cite{mirzaei2023spin,hu2023nerf}. Performing these tasks directly on NeRFs allow a viewpoint agnostic result, which provides richer and more accurate information on the scene. As accurate these methods may be, however, they suffer from two inherent drawbacks; they can mostly work on only one NeRF method, or can perform only one task that they are designed/trained to do. A desirable way forward is to lift these restrictions, where any task can be applied on any NeRF method. The lifting of these restrictions will create a flexible framework, which has the ability of scaling with any future NeRF method or any task.

In this work, we aim to lift these restrictions by going back to 2D in task execution, while still using NeRFs for 3D scene representation. Staying in 2D in task execution will facilitate NeRF method agnosticity, whereas using NeRFs for 3D scene representation allows us to \textit{explore} the scene to find the areas that adhere to task requirements. Within this setting, we introduce the scene exploration framework, which aims to find the camera poses and viewing angles (\textit{e.g.} NeRF inputs) that would render the views that will adhere to a selected criteria provided by the selected task (see Figure \ref{fig:teaser}). A good practical example would be \textit{Finding Waldo}; imagine you have a NeRF model that represents a scene sufficiently well, and the scene has Waldo somewhere in it. The scene exploration framework, within this context, would aim to find the camera pose/viewing angles which would render an image with Waldo in it, even though he might not be fully visible in any available training image.

To the best of our knowledge, scene exploration has not been formally defined or investigated before in the literature, which means there are no suitable baseline methods to compare against. Therefore, we first propose two baseline methods; guided random search (GRS) and pose interpolation-based search (PIBS). We then cast scene exploration as an optimization problem for efficient and accurate solutions, and propose evolution-guided pose search (EGPS) approach to tackle the problem. Unlike some works which tackle somewhat similar issues \cite{yen2021inerf,adamkiewicz2022vision}, EGPS is task-agnostic, accurate and efficient. We conduct a wide range of experiments using our methods with various criteria for practical use cases, such as photo-composition improvement, saliency maximization and image quality maximization. Our results show that efficient exploration is feasible, and we also show that there is a large room for improvement, which will hopefully spark further research in the field.

Our main contributions can be summarized as follows.

\begin{itemize}
\vspace{-0.5mm}
  \setlength{\itemsep}{1pt}
    \item We formally propose the scene exploration framework, which is a NeRF-agnostic framework that aims to find desired camera poses in a scene space encoded by a NeRF method. 
    \item Due to the lack of methods suitable for testing in our framework, we propose the baseline methods guided random search and pose interpolation-based search.
    \item We then view the scene exploration framework from the lens of constrained optimization, and then propose Evolution-Guided Pose Search method, which is an accurate, efficient and task/NeRF-agnostic method that addresses the scene exploration problem.
    \item Finally, we perform extensive experiments using various tasks within our framework, and show that Evolution-Guided Pose Search performs favourably compared to the baselines.
\end{itemize}

\vspace{-2mm}

\section{Related Work} \label{related_work}

\noindent \textbf{Representing 3D Scenes.} In addition to the earlier, hand-crafted ways of representing scenes in 3D \cite{cyganek2011introduction}, more recent neural rendering methods, such as NeRFs, encode the geometry and appearance information of the scene into the parameters of a neural network, using a collection of 2D photos of the scene \cite{mildenhall2021nerf}. NeRFs have become popular especially due to their success in novel-view synthesis, and there has been numerous follow-up works which have improved their accuracy \cite{barron2021mip,deng2022depth}, rendering and convergence speed \cite{muller2022instant,reiser2021kilonerf} and few-shot capabilities \cite{yang2023freenerf,jain2021putting}.

\noindent \textbf{NeRF Use Cases.} Numerous 2D vision methods have been adapted to work on NeRF representations instead of pixels, such as inpainting \cite{mirzaei2023spin}, object detection \cite{hu2023nerf}, segmentation \cite{fan2022nerf} and stylization \cite{zhang2022arf}. Although these results manage to perform the relevant tasks in 3D, and thus lead to improved results compared to their 2D counterparts (especially in terms of spatial consistency), they are inherently limited in two aspects; first, they are highly specialized in terms of the NeRF method they work, which means it is non-trivial to transfer them to another NeRF method, if possible at all. Second, and most importantly, they are specifically designed for a single task (e.g. inpainting, segmentation), and therefore can not perform at all in other tasks. We identify these two aspects as the primary shortcomings of existing use cases for NeRFs, and aim to fix these with our scene exploration framework. 

\noindent \textbf{Navigating scene spaces.} 3D photography methods render the same scene from different views to create animations, but most methods use fixed pose trajectories and thus do not search the scene space to find the desired poses \cite{kopf2020one,shih20203d,fan2023drag}, or vastly limit the search space (\textit{e.g.} a linear trajectory) \cite{googleBlog}. None of these methods use NeRFs as the underlying 3D representation, therefore can not be trivially repurposed to work in our framework. In \cite{yen2021inerf}, authors propose a gradient-based optimization method to estimate the camera pose of a given image in a scene encoded by NeRF. Another work \cite{adamkiewicz2022vision} proposes a gradient-based optimization to find a trajectory in a space defined by a NeRF for robot navigation, which is performed by finding the trajectory with the lowest volume density to avoid collisions. These methods are not suitable for our framework as their gradient-based optimization nature rules out a vast number of criteria they can be applied to. Furthermore, they are not proposed to address our issue and thus are likely to require fundamental and complex changes to function adequately, which is not within our scope.

\section{Exploring NeRF Scene Spaces} \label{sec:method}
In this section, we formally define the scene exploration framework, highlight its practical use cases, and then present three methods - two baselines and our evolution-guided pose search - to operate within the scene exploration for various criteria.  

\subsection{Preliminaries}
 
NeRF methods estimate a 5D function: $f : (x,y,z, \theta, \phi) \rightarrow (\textbf{c}, \sigma)$, where $(x,y,z)$ is the camera pose, $(\theta, \phi)$ is the viewing angle, $\textbf{c}$ is the color and $\sigma$ is the volume density. NeRF methods operate within the volume rendering pipeline, where a ray is shot for each pixel and NeRF method is queried for sample points along each ray. For each ray, ray-wise integration is performed which produces the final color value for each pixel. Using a set of images of the same static scene, NeRF learns to estimate this 5D function via the reconstruction loss between the predicted images and the ground-truth images. NeRF methods can be parameterized by various building blocks; e.g. MLPs \cite{mildenhall2021nerf}, grids \cite{muller2022instant} and gaussians \cite{kerbl20233d}.

\subsection{Scene Exploration Framework} \label{sec:definition}

\noindent \textbf{Definition.} Given a scene sufficiently well-encoded with a NeRF, we can choose any 5D camera pose as input, and NeRF will generate an image of the scene from the input camera pose. These views will be of the same scene, but they will have different appearance, and even geometry/content (see Figure \ref{fig:teaser} for examples).

Let $f(\cdot;\zeta)$ be the NeRF model (parameterized by $\zeta$) that models a scene and $(x,y,z, \theta, \phi)$ a 5D input pose to the NeRF model. Once queried with this input pose \footnote{Equation \ref{eq:first} omits queries for each sample points and all rays for brevity.}, we get

\begin{equation} \label{eq:first}
\begin{aligned}
\mathbb{I} = f((x,y,z, \theta, \phi); \zeta)
\end{aligned}
\end{equation}

where $\mathbb{I}$ is the rendered image. Assume $\mathbb{I}$ has an appearance $\mathbb{A_{I}}$ and geometry/content $\mathbb{G_{I}}$, estimated by functions $A(\cdot)$ and $G(\cdot)$. We formally define the scene exploration framework as

\begin{equation} \label{eq:second}
\begin{aligned}
find \quad & (\hat{x},\hat{y},\hat{z}, \hat{\theta}, \hat{\phi}) \\
\textrm{s.t.} \quad &  \nabla\mathbb{A} < \epsilon_{A}, \quad \nabla\mathbb{G} < \epsilon_{G}    \\
\end{aligned}
\end{equation}
where $\epsilon_{A}$ and $\epsilon_{G}$ indicate small constants, and $\nabla$ indicates the difference between the appearance and geometry of image $\mathbb{I}$ and the desired appearance $\mathbb{A_{T}}$ and geometry $\mathbb{G_{T}}$. In plain words, the scene exploration framework aims to find camera pose(s) that generate an image with desired appearance and geometry. Note that one can use any $f(\cdot;\zeta)$, $A(\cdot)$ and $G(\cdot)$ in our formulation; \textit{e.g.} $A(\cdot)$ and $G(\cdot)$ might be face detectors, saliency segmentation models, image aesthetics networks, image quality assessment models, object detector/segmentors and many others, whereas $f(\cdot;\zeta)$ can be estimated by any NeRF method. Also note that $(\hat{x},\hat{y},\hat{z}, \hat{\theta}, \hat{\phi})$ can be the pose of an existing image used to train $f(\cdot;\zeta)$, or a novel view. 

\noindent \textbf{Practical Scenarios.} Our formulation lends itself to numerous practical scenarios; \textit{e.g.} finding the viewpoint that maximizes the number of faces, aesthetic quality, or finding the viewpoint that includes or excludes an object (\textit{e.g.} finding Waldo). It is easy to see many real-life applications that can stem from our framework; optimally inserting an object within a scene, automatic generation of the best (\textit{e.g.} defined by image quality assessment) possible image of an augmented-reality environment, interactive hide and seek games in VR, teaching of photography rules and many more.

\subsection{Baselines}
The scene exploration framework has not been formally defined nor addressed before, which means there are no baseline methods. To this end, we introduce two naive baselines; guided random search and pose interpolation-based search.  

\algrenewcomment[1]{\(\triangleright\) #1}
\definecolor{ColorGRS}{rgb}{0.9, 0.72, 0.23}
\definecolor{ColorPBIS}{rgb}{0.24, 0.71, 0.54}
\definecolor{ColorOurs}{rgb}{0.0, 0.72, 0.92}

\begin{algorithm}[!t]
\footnotesize
\caption{\textcolor{ColorGRS}{GRS}, \textcolor{ColorPBIS}{PIBS} baselines and \textcolor{ColorOurs}{EGPS} method.}\label{algo:baselines}
\begin{algorithmic}[1]
\Function{ExploreScene}{$I, C, NE, CC, topk, topc,  f, ME, mode$}
    
\Comment{\textcolor{lightgray}{IN: Training frames: I, poses: C, \# of epochs: NE, f($\cdot$): NeRF model.}} 
    
\Comment{\textcolor{lightgray}{IN: Score estimator: SE($\cdot$), CG: \# of children generated per epoch.}}  
    
\Comment{\textcolor{lightgray}{IN: topk: \# of best elements to return.}}  

\Comment{\textcolor{lightgray}{IN: topc: \# of best elements to consider in each search epoch.}}
    
\Comment{\textcolor{lightgray}{IN: mode: The algorithm mode, can be GRS, PIBS or EGPS.}}   \\ 

\State \textcolor{lightgray}{$\triangleright$Initialization stage.} 
\State population = (I, C) 

\Comment{\textcolor{lightgray}{Candidate list, where each item has an image I and a pose C. }}
\For{${cand \in\{population}\}$}
    \State cand.$\lambda$ = SE(cand.I) \Comment{\textcolor{lightgray}{Compute scores $\lambda$  for initial population.}}
\EndFor
\If{$\textcolor{ColorGRS}{mode==GRS}$}
    \State \textcolor{ColorGRS}{min\_pos, max\_pos = min\_pos(population.C), max\_pos(population.C)}
    \State  \textcolor{ColorGRS}{all\_lookAt = [pose.lookAt \textbf{for} pose \textbf{in} population.C]}
    \State  \textcolor{ColorGRS}{up\_Vec = mean([pose.upVec \textbf{for} pose \textbf{in} population.C])}

    \Comment{\textcolor{lightgray}{Find min and max of train poses. Populate lookAt and upVec lists.}}
\EndIf
\If{$\textcolor{ColorPBIS}{mode==PIBS}$}
    \State \textcolor{ColorPBIS}{explored\_pairs = []}  \Comment{\textcolor{lightgray}{Keeping track of used pairs.}}
\EndIf
\\
\State \textcolor{lightgray}{$\triangleright$Iterative search stage.} 
\For{${epoch \in\{1,...,NE}\}$}
    \State new\_poses = []
    \If{$\textcolor{ColorPBIS}{mode==PIBS}$}
        \State \textcolor{ColorPBIS}{population = sort(population, key="$\lambda$", reverse=True)}
        \State \textcolor{ColorPBIS}{interpolation\_pairs = generate\_pairs(top\_c, explored\_pairs})
    
        \Comment{\textcolor{lightgray}{Sample previously unsampled pose pairs from \textit{topc} poses.}}

        \State \textcolor{ColorPBIS}{interpolation\_step = round(CG / len(interpolation\_pairs))}
    
        \Comment{\textcolor{lightgray}{Divide CG by the number of pairs to find interpolation step.}}

        \For{${pair \in\{1,...,interpolation\_pairs}\}$}
            \State \textcolor{ColorPBIS}{pose1, pose2 = pair[0], pair[1]}

            \State \textcolor{ColorPBIS}{explored\_pairs.append((pose1, pose2))}
            \State \textcolor{ColorPBIS}{ explored\_pairs.append((pose2, pose1))}
            
            \Comment{\textcolor{lightgray}{Update explored pairs list to avoid resampling the same pair.}}
            \State \textcolor{ColorPBIS}{poses = interpolate\_poses(pose1, pose2, interpolation\_step) }
            
            \Comment{\textcolor{lightgray}{Generate interpolation\_step number of new poses for each pair.}}
            \State \textcolor{ColorPBIS}{new\_poses.extend(poses)}
        \EndFor
    \EndIf
    \If{$\textcolor{ColorGRS}{mode==GRS}$}
        \For{${cc \in\{1,...,CG}\}$}
            \State \textcolor{ColorGRS}{origin = rand.sample(min\_pos, max\_pos)}
            \State \textcolor{ColorGRS}{lookAt = rand.sample(all\_lookAt)}
            \State \textcolor{ColorGRS}{new\_pose = generate\_pose(origin, lookAt, up\_Vec)}
            
            \Comment{\textcolor{lightgray}{Sample poses between min and max, and a lookAt vector.}}
            \State \textcolor{ColorGRS}{new\_poses.append(new\_pose)}
        \EndFor
    \EndIf
    \If{$\textcolor{ColorOurs}{mode==EGPS}$}
        \State \textcolor{ColorOurs}{population = sort(population, key="$\lambda$", reverse=True)}
        \For{${cc \in\{1,...,CG}\}$}
            \State \textcolor{ColorOurs}{parent1,parent2=sample\_pair\_biased(population[:middle])}
            
            \Comment{\textcolor{lightgray}{Choose a pair; high $\lambda$ poses are more likely to be chosen.}}
            \State \textcolor{ColorOurs}{origin, lookAt, up\_Vec = crossover(parent1, parent2)}
            \State \textcolor{ColorOurs}{origin, lookAt = mutation(origin, lookAt)  }                  
            \State \textcolor{ColorOurs}{new\_pose = generate\_pose(origin, lookAt, up\_Vec)}
            \State \textcolor{ColorOurs}{new\_poses.append(new\_pose)}
        \EndFor
    \EndIf
    
    \For{${new\_pose \in\{new\_poses}\}$}
        \State new\_img = f(new\_pose) \Comment{\textcolor{lightgray}{Render image.}}
        \State new\_score = SE(new\_img) \Comment{\textcolor{lightgray}{Compute scores.}}
        \State population.append((new\_img, new\_pose, new\_score))
    \EndFor
\EndFor
\\
\State \textcolor{lightgray}{$\triangleright$Output.} 
\State sorted\_population = sort(population, key="$\lambda$", reverse=True)
\State \textbf{return} sorted\_population[:topk])       \Comment{\textcolor{lightgray}{OUT: Best top-k candidates.}}
\EndFunction
\end{algorithmic}

\end{algorithm}

\noindent \textbf{Guided random search (GRS).} There are too many locations one can place cameras in any scene, and many viewing angles, and these locations increase dramatically when scene scale is large. Therefore, a true random search would be infeasible. To this end, we propose guided-random search, where we constrain the random placement of new camera poses close to the camera poses of training images, since most $f(\cdot;\zeta)$ formulations perform better near training poses. See Algorithm \ref{algo:baselines} for a breakdown on \textcolor{ColorGRS}{GRS}.

GRS operates in an iterative framework, where in the initialization stage we compute the score \footnote{The word \textit{score} is used to denote the output of $\mathbb{A}$ and $\mathbb{G}$ throughout the paper. See Sections 4.2, 4.3 and 4.4 for use-case specific details.} for each available image and find the min and max coordinates among available images. In the iterative search stage, we randomly sample new coordinates within min and max value, while sampling lookAt vectors from available posed images. Using the mean of existing up vectors, the sampled coordinates and lookAt vectors, we create our new camera poses. New images are rendered from new camera poses using $f(\cdot;\zeta)$, and scores are computed. This process is repeated until the epoch limit is reached, where each epoch adds new poses to the population. The camera poses that best satisfy Equation \ref{eq:second} are then chosen.

\noindent \textbf{Pose interpolation-based search (PIBS).} PIBS, similar to GRS, operates in an iterative framework. Unlike GRS, however, PIBS removes the randomness in pose generation, but introduces another layer of randomness in choosing pose pairs between which we generate new poses. See Algorithm \ref{algo:baselines} for a breakdown of \textcolor{ColorPBIS}{PIBS}.

In each epoch, we sample all available pairs of poses among the poses with high scores, and simply generate new poses between the chosen pair via SLERP \cite{shoemake1985animating} that performs a rotation with a constant angular velocity. We keep track of the chosen pairs so as not to sample them again in the next epochs. This process is repeated until we reach the epoch limit, where each epoch adds new camera poses to the population. Similar to GRS, the camera poses that best satisfy Equation \ref{eq:second} are chosen among the final population. PIBS is not as diverse as GRS in the poses it generates, but it manages to uniformly cover the regions of interest thanks to the pose interpolation.

\subsection{Evolution-Guided Pose Search}
Our scene exploration framework, as Equation \ref{eq:second} shows, is effectively a constrained optimization problem. GRS and PIBS baselines are naive baseline methods that do not perform any optimization, and they are not concerned to be efficient in coming up with a solution. Unlike some methods addressing similar issues to ours \cite{adamkiewicz2022vision,yen2021inerf}, we aim to accommodate all types of $A(\cdot)$ and $G(\cdot)$, which requires agnosticity to several factors, such as search criteria differentiability and search space convexity. To this end, we propose Evolution-Guided Pose Search (EGPS) that is based on the genetic algorithm \cite{holland1992adaptation}, which accommodates for all these factors while being accurate and cheaper compared to alternatives (e.g. reinforcement learning). See Algorithm \ref{algo:baselines} for a breakdown of \textcolor{ColorOurs}{EGPS}.

In each epoch, EGPS first samples two parent poses among the \textit{better} half of the population (\textit{e.g.} the population with high scores) with a skewed distribution, where higher scores means higher chance of selection. We then perform the \textit{crossover} operation on the parent poses $\mathbb{P}_{1}$ and $\mathbb{P}_{2}$, which gives us

\begin{equation} \label{eq:crossover}
\begin{aligned}
upVec_x = (\mathbb{P}_{{1}_{upVec_{x}}} + \mathbb{P}_{{2}_{upVec_{x}}}) / 2  \\
upVec_y = (\mathbb{P}_{{1}_{upVec_{y}}} + \mathbb{P}_{{2}_{upVec_{y}}}) / 2  \\
upVec_z = (\mathbb{P}_{{1}_{upVec_{z}}} + \mathbb{P}_{{2}_{upVec_{z}}}) / 2  \\
origin_x \sim \mathcal{U}([\mathbb{P}_{1}.origin_{x}, \mathbb{P}_{2}.origin_{x}])  \\
origin_y \sim \mathcal{U}([\mathbb{P}_{1}.origin_{y}, \mathbb{P}_{2}.origin_{y}])  \\
origin_z \sim \mathcal{U}([\mathbb{P}_{1}.origin_{z}, \mathbb{P}_{2}.origin_{z}])  \\
lookAt_x \sim \mathcal{U}([\mathbb{P}_{1}.lookAt_{x}, \mathbb{P}_{2}.lookAt_{x}])  \\
lookAt_y \sim \mathcal{U}([\mathbb{P}_{1}.lookAt_{y}, \mathbb{P}_{2}.lookAt_{y}])  \\
lookAt_z \sim \mathcal{U}([\mathbb{P}_{1}.lookAt_{z}, \mathbb{P}_{2}.lookAt_{z}])  \\
\end{aligned}
\end{equation}
\normalsize

where $\mathcal{U}$ denotes a uniform distribution, $upVec$ indicates the up vector, $origin$ indicates the coordinates and $lookAt$ indicates the lookAt vector. 

After obtaining the post-crossover components, we apply a round of \textit{mutation} on lookAt and origin components. The strength of the mutation is controlled with $\nabla$ values, which are defined for each component as 

\begin{equation} \label{eq:nablas}
\begin{aligned}
\nabla_{origin_{x}} = 0.1 * (\mathbb{P}_{1}.origin_{x} - \mathbb{P}_{2}.origin_{x}) \\
\nabla_{origin_{y}} = 0.1 * (\mathbb{P}_{1}.origin_{y} - \mathbb{P}_{2}.origin_{y}) \\
\nabla_{origin_{z}} = 0.1 * (\mathbb{P}_{1}.origin_{z} - \mathbb{P}_{2}.origin_{z}) \\
\nabla_{lookAt_{x}} = 0.1 * (\mathbb{P}_{1}.lookAt_{x} - \mathbb{P}_{2}.lookAt_{x}) \\
\nabla_{lookAt_{y}} = 0.1 * (\mathbb{P}_{1}.lookAt_{y} - \mathbb{P}_{2}.lookAt_{y}) \\
\nabla_{lookAt_{z}} = 0.1 * (\mathbb{P}_{1}.lookAt_{z} - \mathbb{P}_{2}.lookAt_{z}) \\
\end{aligned}
\end{equation}
\normalsize

Using the $\nabla$ values, we then perform the mutation operation, which we define as

\begin{equation} \label{eq:mutation}
\begin{aligned}
origin_x = origin_x + \delta_{origin_{x}} \\
origin_y = origin_y + \delta_{origin_{y}}  \\
origin_z = origin_z + \delta_{origin_{z}}  \\
lookAt_x = lookAt_x + \delta_{lookAt_{x}} \\
lookAt_y = lookAt_y + \delta_{lookAt_{y}} \\
lookAt_z = lookAt_z + \delta_{lookAt_{z}} \\
\end{aligned}
\end{equation}
\normalsize

where the update values $\delta$ are sampled as

\begin{equation} \label{eq:mutation}
\begin{aligned}
\delta_{origin_{x}} \sim  PMF(-\nabla_{origin_{x}}, 0, \nabla_{origin_{x}};0.05,0.9,0.05) \\
\delta_{origin_{y}} \sim  PMF(-\nabla_{origin_{y}}, 0, \nabla_{origin_{y}};0.05,0.9,0.05) \\
\delta_{origin_{z}} \sim  PMF(-\nabla_{origin_{z}}, 0, \nabla_{origin_{z}};0.05,0.9,0.05) \\
\delta_{lookAt_{x}} \sim  PMF(-\nabla_{lookAt_{x}}, 0, \nabla_{lookAt_{x}};0.05,0.9,0.05) \\
\delta_{lookAt_{y}} \sim  PMF(-\nabla_{lookAt_{y}}, 0, \nabla_{lookAt_{y}};0.05,0.9,0.05) \\
\delta_{lookAt_{z}} \sim  PMF(-\nabla_{lookAt_{z}}, 0, \nabla_{lookAt_{z}};0.05,0.9,0.05) \\
\end{aligned}
\end{equation}
\normalsize

where PMF indicates the probability mass function with given values and their respective probabilities. Note that the probability distribution centres on 0, which means we mostly refrain from adding any perturbation during the mutation operation. Furthermore, the added perturbation is set as 10\%  of the distance of the parents, which is another limiting factor to avoid divergence during the exploration process. The resulting components $upVec$, $lookAt$ and $origin$ are used to generate new poses at each epoch. Similar to GRS and PIBS, the camera poses that best satisfy Equation \ref{eq:second} are chosen among the final population. EGPS is inherently random, but centres its randomness on successful parents, which provides a good trade-off between diversity and accuracy, as we will show in the following sections. 
\section{Experiments} \label{experiments}
In this section, we first describe our experimental setup, including the NeRF method we use, the metrics used to measure the success of scene exploration methods and datasets experimented on. We then present our results and discussions with several practical use cases/criteria within the scene exploration framework. We analyse our results and finalize with a discussion on current progress and limitations. 

\subsection{Experimental Setup}
\noindent \textbf{NeRF method.} As noted in the previous sections, our framework can make use of any NeRF method. Throughout the paper, however, we use Instant-NGP \cite{muller2022instant} as the underlying NeRF method to encode/represent a scene. We choose Instant-NGP due to several reasons; it i) converges in a matter of minutes, thus enabling fast experimentation, ii) performs adequately in real-life scenes, which we base our experimentations on and iii) renders in real-time, which would be a prime requirement for practical, deployment use cases. Briefly, Instant-NGP uses implicit representations in the form of a multi-resolution hash grid for fast memory access. It removes some capacity from density and color MLPs and instead uses features placed on a voxel grid, where the coordinates of a ray sample point is used to interpolate the features of the corresponding voxel edges to produce a feature map. This process is performed for each resolution, and generated features are concatenated and fed to the density MLP to generate the volume density for the given coordinate. This density output is then used with encoded viewing angle inputs to generate the final color value through the color MLP. We train Instant-NGP on each scene using the original codebase with the original training settings.  

\noindent \textbf{Datasets.} We envisage the scene exploration tasks to operate in real-life scenarios, which largely eliminate the value of synthetic datasets in our context. Therefore, we experiment exclusively on real-life datasets. To capture the in-the-wild nature of our problem, we sample scenes from several datasets; \textit{horns} and \textit{trex} from LLFF \cite{mildenhall2019local}, \textit{face} from Instruct-NeRF2NeRF \cite{haque2023instruct}, \textit{fox} from Instant-NGP \cite{muller2022instant}, \textit{family} from Tanks and Temples \cite{knapitsch2017tanks} and a new scene we collected named \textit{plant\_new}. Except \textit{family}, most of the scenes are front-facing with a prominent object in the scene center, whereas \textit{family} includes images from the whole surrounding of the primary scene object.  We train Instant-NGP on these scenes using only the training images, which range from 35 (\textit{plant\_new}) to 153 (\textit{family}). We use the provided camera poses when available, or generate poses with COLMAP \cite{schonberger2016structure} ourselves if not.

\noindent \textbf{Evaluation Metrics.} The primary aim of the scene exploration framework is to find views that adhere to some criteria. As discussed in Section \ref{sec:definition}, these views can be already present in the training images, or can be novel views unseen during training. Motivated by this, we propose to quantify the increase in the selected criteria; essentially, we want to see if the search method managed to generate a novel view that is better than any training image, and if it did, we want to measure by how much. An example would be finding Waldo; \textit{e.g.} training images might have Waldo in them partially, but if the search method can find a pose that generates a novel view with Waldo fully in it, we call it a success and try to measure by how much Waldo is more visible compared to the training image that had the \textit{best visible} Waldo.  

The scene exploration framework can house many different criteria, as discussed previously. Therefore, it is not a surprise to see that various criteria will have different value scales; \textit{e.g.} saliency maximization will count the number of pixels and can easily go up to hundreds of thousands, photo-composition scoring networks will generate a singular score between 1 and 5, no-reference image quality assessment scores will be between 1 and 100, and so on. This discrepancy in value scales requires a normalization, which we perform by simply measuring the \% improvement in criteria value. First, we compute this metric for the top value; \textit{e.g.} the highest criteria value yielded by the search method. We call this metric \textit{Criteria Value Improvement Ratio} (CVIR), and define it as 

\begin{equation}
    CVIR= \left(\frac{\Gamma(\mathbb{AG}(I_{all})}{\Gamma(\mathbb{AG}(I{_{tr}}))} - 1\right) * 100
\end{equation}

where $\Gamma$ is an operator that finds the maximum or minimum of a set of values depending on the chosen criteria, $\mathbb{AG}$ is the function that generates the criteria value (\textit{e.g.} a neural net that provides photo-composition scores) \footnote{In Equation \ref{eq:second}, we use $A(\cdot)$ and $G(\cdot)$ to estimate appearance and geometry. Here, we assume a single function $\mathbb{AG}$ for both (or only one, depending on the use case) for brevity. For other use cases, separate functions $A(\cdot)$ and $G(\cdot)$ might be needed.}, $I_{tr}$ is the set of training images and $I_{all}$ is the union of training images and novel images generated by the search method. 
\begin{table*}[ht!]
\resizebox{0.9\textwidth}{!}{%
\begin{tabular}{l|cc|cc|cc|cc|cc|cc}
\multicolumn{1}{c|}{} & \multicolumn{2}{c|}{trex}   & \multicolumn{2}{c|}{horns}  & \multicolumn{2}{c|}{fox}      & \multicolumn{2}{c|}{face} & \multicolumn{2}{c|}{family}  & \multicolumn{2}{c}{plant\_new} \\ \hline
\multicolumn{1}{c|}{} & CVIR         & mCVIR        & CVIR         & mCVIR        & CVIR          & mCVIR         & CVIR    & mCVIR           & CVIR         & mCVIR         & CVIR           & mCVIR         \\ \hline
GRS                   & 1.8          & 2.3          & 0.7          & 2.6          & 16.0          & 10.0          & 0.0     & \textbf{0.3}    & 0.0          & 1.0           & 0.0              & 2.5           \\
PIBS                  & 1.5          & 2.7          & \textbf{3.2} & \textbf{6.9} & 19.6          & 23.2          & 0.0     & 0.0             & \textbf{3.6} & \textbf{9.9}  & 0.0            & \textbf{8.6}  \\
EGPS                  & \textbf{3.6} & \textbf{4.1} & 3.1          & 6.3          & \textbf{31.1} & \textbf{29.0} & 0.0     & 0.1             & 0.0            & 1.2           & 0.0            & 3.5           \\ \hline
GRS $\dagger$                 & 2.5          & 3.1          & 3.1          & 5.2          & 28.5          & 27.0          & 0.0     & 1.8             & \textbf{3.7} & \textbf{10.0} & 3.6            & 9.0           \\
PIBS $\dagger$                & 1.9          & 3.2          & 4.3          & 8.1          & 15.7          & 18.8          & 0.0     & 0.9             & \textbf{3.7} & \textbf{10.0} & 0.0            & 12.0          \\
EGPS $\dagger$                & \textbf{3.7} & \textbf{4.7} & \textbf{5.6} & \textbf{8.6} & \textbf{37.3} & \textbf{38.5} & 0.0     & \textbf{4.9}    & 1.76         & 5.0           & \textbf{11.5}  & \textbf{23.0} \\ \hline
\end{tabular}
}
    \caption{Scene exploration framework, with photo-composition score improvement as the search criteria. The results are given for all search methods, reported as CVIR and mCVIR metrics. \textit{The higher the better.} Rows indicated with $\dagger$ are evaluations in \textit{high-pose} regime, and the others are in \textit{low-pose} regime.}
    \label{tab:aesthetics}
\end{table*}

\begin{figure*}[!ht]
  \centering
        \includegraphics[width=1.0\textwidth]{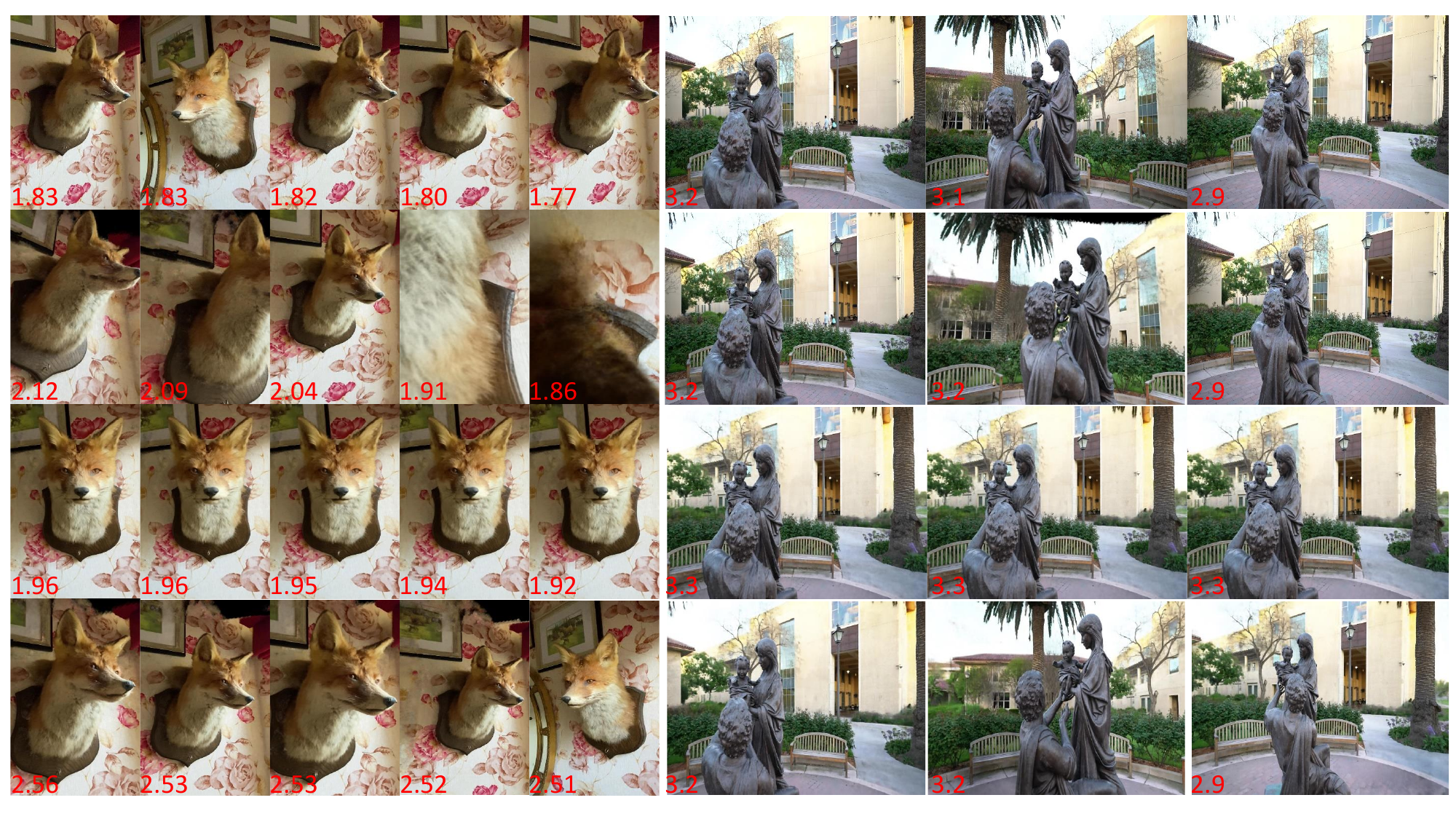}
        \vspace{-4mm}
  \caption{Qualitative examples for each search method for the scenes \textit{fox} and \textit{family} for photo-composition improvement. For each scene, top row shows the best training images, second row shows the best results of GRS, third row shows the best results of PIBS and the last row shows the results of EGPS. Overlaid on images are scores generated with \cite{zhang2021image}. } 
  \label{fig:aesthetics}
  \vspace{-2mm}
\end{figure*}

\begin{table*}[ht!]
\resizebox{0.9\textwidth}{!}{%
\begin{tabular}{l|cc|cc|cc|cc|cc|cc}
\multicolumn{1}{c|}{} & \multicolumn{2}{c|}{trex} & \multicolumn{2}{c|}{horns}  & \multicolumn{2}{c|}{fox}      & \multicolumn{2}{c|}{face} & \multicolumn{2}{c|}{family} & \multicolumn{2}{c}{plant\_new} \\ \hline
\multicolumn{1}{c|}{} & CVIR    & mCVIR           & CVIR         & mCVIR        & CVIR          & mCVIR         & CVIR        & mCVIR       & CVIR         & mCVIR        & CVIR          & mCVIR          \\ \hline
GRS                   & 0.0     & 0.2             & 0.3          & 0.4          & 61.0          & 68.5          & 0.0         & 0.0         & 0.4          & 0.4          & 0.0           & 0.0            \\
PIBS                  & 0.0     & \textbf{0.7}             & 0.0          & 0.8          & 54.2          & 65.7          & 0.0         & 0.0         & \textbf{4.8} & \textbf{6.4} & 0.0           & 0.0            \\
EGPS                  & 0.0     & 0.5             & \textbf{2.6} & \textbf{1.6} & \textbf{65.3} & \textbf{76.8} & 0.0         & 0.0         & 2.5          & 3.3          & 0.0           & 0.0            \\ \hline
GRS $\dagger$                 & 0.0     & 0.5             & 0.1          & 1.10         & 64.8          & 72.7          & 0.0         & 0.0         & 3.3          & 3.2          & 0.0           & 0.0            \\
PIBS $\dagger$                & 0.0     & 0.6             & 0.4          & 1.1          & 57.7          & 68.6          & 0.0         & 0.0         & \textbf{6.5} & 7.1          & 0.0           & 0.0            \\
EGPS $\dagger$                & 0.0     & \textbf{0.8}    & \textbf{3.6} & \textbf{4.4} & \textbf{70.0} & \textbf{81.3} & 0.0         & 0.0         & 5.6          & \textbf{7.2} & 0.0           & 0.0            \\ \hline
\end{tabular}
}
    \caption{Scene exploration framework, with no-reference image quality assessment score improvement as the search criteria. The results are given for all search methods, reported as CVIR and mCVIR metrics. \textit{The higher the better.} Rows indicated with $\dagger$ are evaluations in \textit{high-pose} regime, and the others are in \textit{low-pose} regime.}
    \label{tab:iqa}
\end{table*}

\begin{figure*}[!ht]
  \centering
        \includegraphics[width=1.0\textwidth]{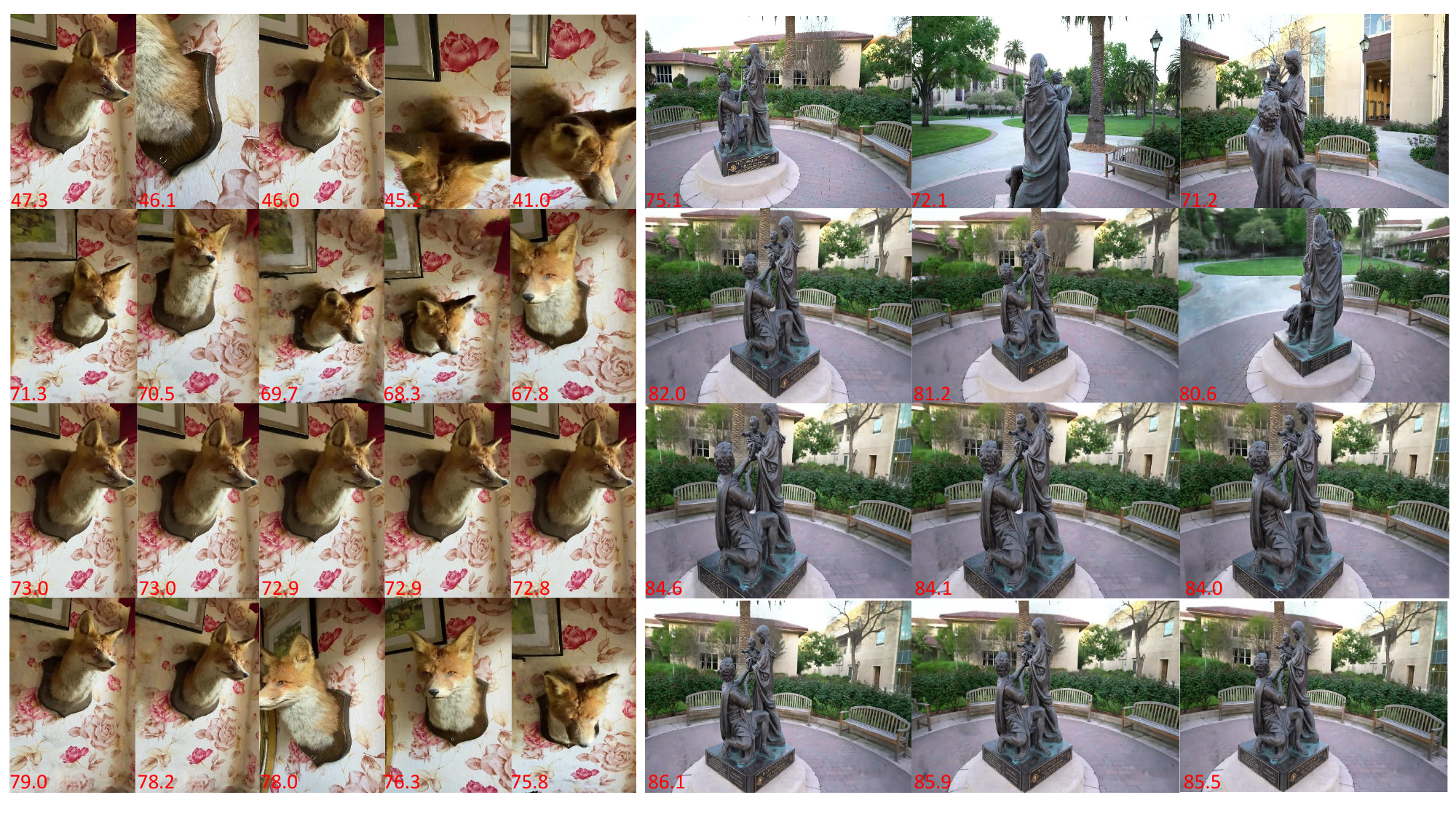}
        \vspace{-4mm}
  \caption{Qualitative examples for each search method for the scenes \textit{fox} and \textit{family} for image quality maximization. For each scene, top row shows the best training images, second row shows the best results of GRS, third row shows the best results of PIBS and the last row shows the results of EGPS. Overlaid on images are their scores generated with \cite{madhusudana2022image}.} 
  \label{fig:iqa}
  \vspace{-2mm}
\end{figure*}

It is possible that the generation of novel views might not lead to an image with the new best criteria value. However, the search method could generate new desirable poses. For accurately quantifying such cases, we propose an extension of CVIR we call \textit{mean} \textit{Criteria Value Improvement Ratio} (mCVIR), which is defined as 

\begin{equation}
    mCVIR= \left(\frac{mean(\Gamma_{N}(\mathbb{AG}(I_{all})))}{mean(\Gamma_{N}(\mathbb{AG}(I{_{tr}})))} - 1\right) * 100
\end{equation}

where $N$ indicates the number of outputs generated by $\Gamma$; \textit{e.g.} instead of taking only the best, we now take 10 best (best defined by maximum or minimum) poses. This simple extension lets us measure the impact of the search method in a fine-grained manner. 

\noindent \textbf{Implementation details.} In our experiments, we use the \textit{maximum} operator for $\Gamma$ and use $N=10$. Furthermore, we compute CVIR and mCVIR in two different regimes; low-pose (LP) and high-pose (HP). In low-pose regime, we evaluate the convergence characteristics of the search methods by restricting the number of poses they can generate in a way that the number of all images (training and generated) is between 250-300. High-pose regime evaluates the full-power of the search method by letting the method generate more and more poses, up to a value between 900 to 1000. These two regimes allow for a fair comparison between different search methods in a fixed-budget manner. To meet these limits, we tune the \textit{children generated per epoch} parameter. All methods are run for 5 epochs with $topk=10$. See Algorithm \ref{algo:baselines} for further details.

\subsection{Photo-composition improvement} \label{exp:aesthetics}
\noindent \textbf{Details.} We first test the baselines and our method in an aesthetics-maximization setting, where the idea is to find novel views which would increase the aesthetics value of an image depicting a given scene. Since aesthetics is a highly subjective criteria, we focus on composition-based aesthetics, where a given frame is evaluated on how it conforms to various photo-composition rules; \textit{e.g.} symmetry, recurrence, rule of thirds, diagonals, etc. We use SAMP-Net \cite{zhang2021image} as $\mathbb{AG}$ in this section. SAMP-Net takes in an image and in one branch, predicts 5 attributes related to following photo-composition rules; rule of thirds, balancing elements, object emphasis, symmetry, and repetition. In another branch, it predicts a distribution of scores for the aesthetics scoring, where the features from the attribute branch is used to influence the final aesthetics score. We use the original SAMP weights \cite{zhang2021image}, which produces an aesthetics score between 1 and 5. The results are shown in Table \ref{tab:aesthetics}.

\noindent \textbf{Results.} We first analyse the low-pose regime results; in two scenes, we see that none of the search methods managed to outperform the training images, which is expected as the newly generated poses are at most in the number of 150. EGPS outperforms others in two of the scenes where all methods manage to improve, and PIBS outperforms in two. In total, PIBS and EGPS are quite close to each other and mostly outperform GRS.

Increasing the number of poses dramatically shift the winds towards EGPS' favour; it manages to clearly outperform others in all but one scene (in mCVIR, see Table \ref{tab:aesthetics}). CVIR and mCVIR values are mostly higher than low-pose results, which is expected as each method has more chance to find a better pose. 

\begin{table*}[t!]
\resizebox{0.9\textwidth}{!}{%
\begin{tabular}{l|cc|cc|cc|cc|cc|cc}
\multicolumn{1}{c|}{} & \multicolumn{2}{c|}{trex}     & \multicolumn{2}{c|}{horns}    & \multicolumn{2}{c|}{fox}      & \multicolumn{2}{c|}{face} & \multicolumn{2}{c|}{family}   & \multicolumn{2}{c}{plant\_new} \\ \hline
\multicolumn{1}{c|}{} & CVIR          & mCVIR         & CVIR          & mCVIR         & CVIR          & mCVIR         & CVIR        & mCVIR       & CVIR          & mCVIR         & CVIR           & mCVIR         \\ \hline
GRS                   & 13.2          & 17.6          & 12.6          & 6.9           & 0.0           & 18.1          & 0.0         & 0.0         & 35.0          & 28.1          & 0.0            & 1.6           \\
PIBS                  & 0.8           & 7.7           & 1.5           & 8.8           & 13.7          & \textbf{53.0} & 0.0         & 0.0         & 56.7          & \textbf{87.9} & \textbf{29.2}  & \textbf{48.8} \\
EGPS                  & \textbf{19.2} & \textbf{25.6} & \textbf{38.1} & \textbf{35.3} & \textbf{14.4} & 49.6          & 0.0         & 0.0         & \textbf{72.5} & 61.1          & 4.1            & 13.2          \\ \hline
GRS $\dagger$                   & 28.7          & 30.9          & 16.2          & 10.9          & 0.0           & 24.3          & 0.0         & 0.0         & 61.3          & 78.2          & 13.1           & 17.8          \\
PIBS $\dagger$                  & 11.9          & 19.4          & 24.8          & 33.1          & 14.1          & 53.2          & 0.0         & 0.0         & 61.5          & 93.7          & 30.4           & 52.8          \\
EGPS $\dagger$                   & \textbf{36.0} & \textbf{42.9} & \textbf{60.8} & \textbf{64.4} & \textbf{19.2} & \textbf{58.7} & 0.0         & 0.0         & \textbf{76.7} & \textbf{93.9} & \textbf{37.4}  & \textbf{57.4} \\ \hline
\end{tabular}
}
    \caption{Scene exploration framework, with salient area maximization as the search criteria. The results are given for all search methods, reported as CVIR and mCVIR metrics. \textit{The higher the better.} Rows indicated with $\dagger$ are evaluations in \textit{high-pose} regime, and the others are in \textit{low-pose} regime.}
    \label{tab:saliency}
\end{table*}

\begin{figure*}[!ht]
  \centering
        \includegraphics[width=1.0\textwidth]{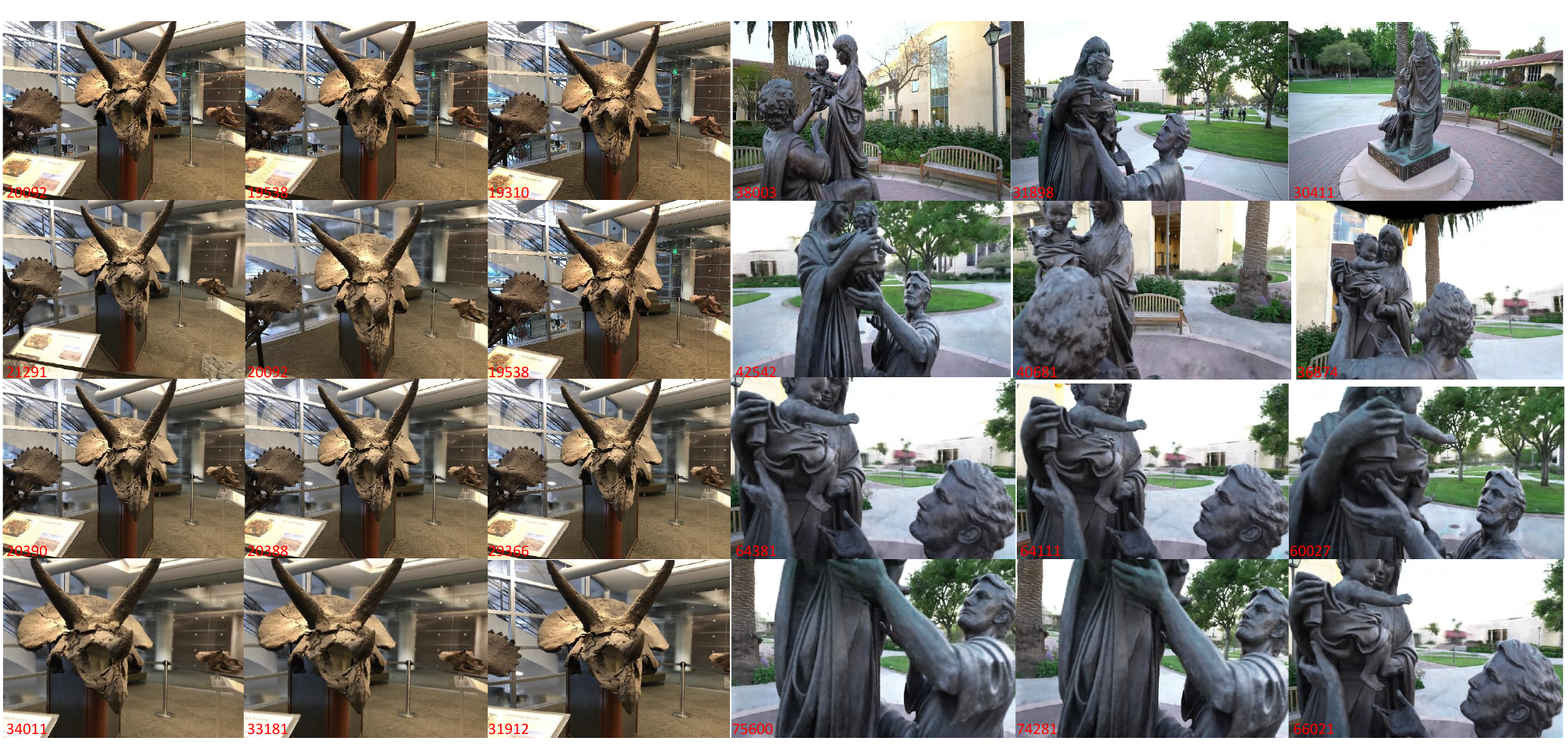}
        \vspace{-4mm}
  \caption{Qualitative examples for each search method for the scenes \textit{horns} and \textit{family} for saliency maximization.  For each scene, top row shows the best training images, second row shows the best results of GRS, third row shows the best results of PIBS and the last row shows the results of EGPS. Overlaid on images are the number of pixels predicted to be salient via \cite{qin2019basnet}. } 
  \label{fig:saliency}
  \vspace{-2mm}
\end{figure*}

\begin{figure*}[!ht]
  \centering
        \includegraphics[width=1.0\textwidth]{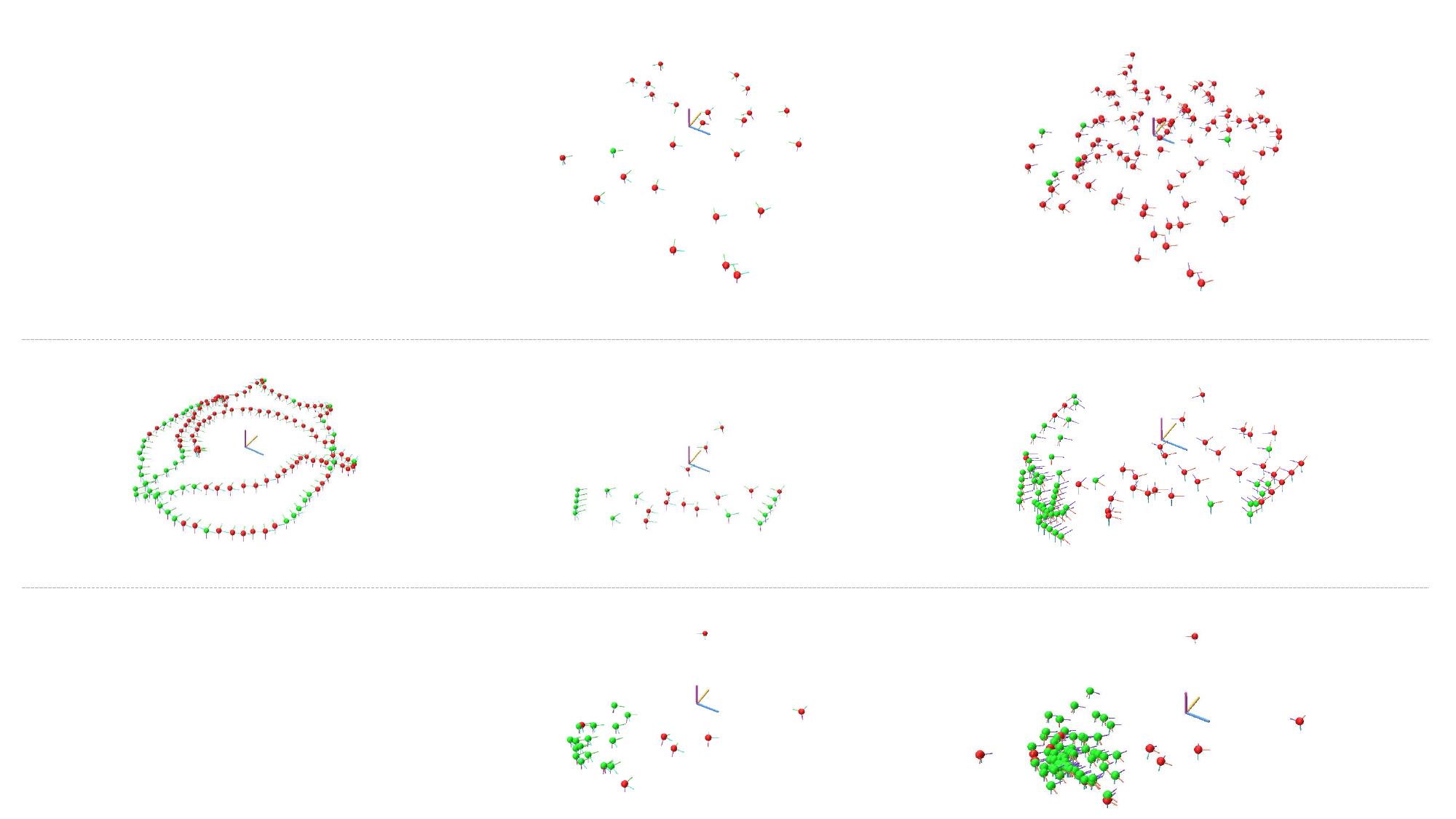}
        \vspace{-4mm}
  \caption{Visualization of the camera poses of the \textit{family} scene, where the scene centre is shown with the axes lines. Each point corresponds to an image pose (training or novel), where \textcolor{green}{green} points are above an IQA score threshold and \textcolor{red}{red} ones are not. Left-most image shows the poses of available images in the beginning, whereas middle column are the results generated in the first epoch, and the right-most column are the poses generated after 5 epochs. From top to bottom, GRS, PIBS and EGPS results are shown. Better viewed when zoomed in.} 
  \label{fig:pointcloud}
  \vspace{-2mm}
\end{figure*}

\subsection{Image Quality Maximization }
\noindent \textbf{Details.} In addition to the photo-composition improvement criteria, we test all three methods within an image quality maximization setting, where the goal is to find novel views which would increase the overall quality value of an image that depicts a given scene. We adopt a no-reference image quality assessment (IQA) method for practical reasons, and choose CONTRIQUE \cite{madhusudana2022image} as our model that acts as the $\mathbb{AG}$. CONTRIQUE relies on self-supervised learning where a model is pretrained to predict a distortion type and intensity for a set of distortions, and this model is then finetuned to predict quality predictions with the addition of a simple regression layer. We use the original CONTRIQUE weights, which produces an image quality score between 0 and 100. The results are shown in Table \ref{tab:iqa}.


\noindent \textbf{Results.} The major difference compared to Table \ref{tab:aesthetics} is that two scenes show absolutely no improvement, either in low or high-pose regimes. We find the reason to be CONTRIQUE; we observe that it does a  good job in detecting NeRF artefacts, despite not being trained on them. This means that although the NeRF models are trained well, one would inevitably encounter artefacts in images rendered from novel views due to inherent seen image bias in NeRF models. This is exactly what happens in the search methods, so in these two scenes training images never fall behind the newly rendered images. Note that the same behaviour is not necessarily captured by SAMPNet in the photo-composition improvement setting, which focuses on a different characteristics of images.

In low-pose regime, ERPS leads the others on all scenes (except \textit{family}) that methods actually manage to work. In high-pose, the same trend continues, where some results (\textit{e.g.} \textit{family} mCVIR) become even better for EGPS. GRS produces the worst results consistently except \textit{fox} scene. In general, we see that high-pose regime results are higher than low-pose regime results, which is expected.

\subsection{Saliency Maximization}
\noindent \textbf{Details.} Our final use case within is the saliency maximization setting, where the goal is to find novel views which would increase the number of pixels predicted to be salient in an image depicting a given scene. We use the popular saliency prediction method BASNet \cite{qin2019basnet} to act as the $\mathbb{AG}$. BASNet proposes a two-stage U-Net architecture where the first stage predicts a coarse saliency map, and the second stage refines this saliency map to produce finer details, and trains this architecture with a novel hybrid loss that comprises cross-entropy, SSIM and IoU losses \cite{skartados2023trickvos}. We use the original BASNet weights trained on the DUTS-TR dataset \cite{wang2017learning}, which produces a binary mask that indicates the salient pixels. We then count the number of pixels predicted as salient, and aim to maximize this number. The results are shown in Table \ref{tab:saliency}.

\noindent \textbf{Results.} We see a clearer picture in saliency maximization; across the board improvements for each method can be observed in most of the scenes. In the low-pose regime, EGPS has a decisive lead, with PIBS leading only on the \textit{plant\_new} scene. In the high-pose regime, EGPS maintains its lead (except the \textit{face} scene), where none of the other methods manage to produce an improvement. We also note that EGPS' lead is much more clear-cut compared to other use cases; it leads in both CVIR and mCVIR metrics. As expected, high-pose results consistently outperform low-pose regime results.

\subsection{Qualitative Results}
\noindent \textbf{Photo-composition improvement.} We provide qualitative results for photo-composition improvement experiments in Figure \ref{fig:aesthetics}. The visuals show that GRS is indeed random and tends to produce diverse results in terms of appearance. PIBS, on the other hand, shows its non-random nature and tends to produce images that are basically the same in appearance, thus lacks diversity. EGPS manages to deliver its promise; being accurate while being diverse. It produces candidates visibly different from each other, while producing images better than the original training images in the \textit{family} scene. Note that there are images with specific NeRF artefacts; \textit{e.g.} floaters and/or zero-density areas. SAMPNet is not designed to detect such failure modes, therefore this behaviour is not unexpected. The visuals largely follow the photo-composition rules; \textit{e.g.} in the fox scene, PIBS tends to find a pose with a high symmetry score, and holds onto it, whereas EGPS produces images that have high object emphasis scores. In the \textit{family} scene, most of the resulting images have high rule-of-thirds score, with some images showing high object emphasis and symmetry score as well.  

\noindent \textbf{Image quality maximization.} The visual results for image quality maximization are shown in Figure \ref{fig:iqa}. Similar to the photo-composition case, we see PIBS having virtually no diversity in its results due to its non-random nature, while GRS and EGPS produces diverse results. However, EGPS provides little diversity on the \textit{family} scene as well. The visuals follow a pattern that is in line with what CONTRIQUE evaluates; in high-scoring images, image quality degradation factors such as blur and white noise tend to be less present. Note that NeRF artefacts are still present in some results, though they are fewer in number as CONTRIQUE tends to take such artefacts in its scoring, at least up to some degree.

\noindent \textbf{Saliency maximization.} The qualitative results for saliency maximization are shown in  Figure \ref{fig:saliency}. We see similar trends to the previous settings, where PIBS producing no diversity while GRS and EGPS producing diverse results, although the diversity is more prominent in the \textit{family} scene compared to the \textit{horns} scene. Possibly the most visually interpretable results are produced in this case; saliency maximization aims to increase the number of salient pixels. The figures show that this translates to a form of zooming-in to the objects. This zooming in effect is significantly more prominent in EGPS, which outperforms the other two in quantitative performance as well. Occasional zero-density areas are present due to the performance of the underlying NeRF representation.

\noindent \textbf{What is actually happening?} We show a 3D visualization of the camera poses in \textit{family} scene to show the behaviour of each search method in Figure \ref{fig:pointcloud} for image quality maximization. The training images are well-rounded, where the images cover most of the scene.

In the first epoch, GRS behaves as expected and covers a wider part of the scene, though it fails to produce poses that have the desired characteristics. PIBS and EGPS produces much more successful poses, where PIBS manages to poses that are spread out. EGPS, on the other hand, has a better success ratio in generating good poses (\textit{e.g.} much fewer red points compared to PIBS).

In the final result, GRS covers the scene more uniformly, but ultimately fails in delivering successful poses. PIBS and EGPS does a much better job in generating better poses. Note that PIBS tends to generate poses that are quite close to initially available images, it does not take any risks. EGPS, on the other hand, finds poses that are more diverse; the size of the points correspond to their elevation in the scene, and in PIBS results, one can see that they have similar sizes, which means they are more or less of the same elevation. EGPS, however, tends to tackle this limitation and produces poses with more diverse coordinates and viewing angles. Note that PIBS produces failure poses that are close to the scene centre, which is a result of picking parents that are opposite each other (taking scene centre as their origin). EGPS avoids that, and is much more successful in terms of generated good to bad poses ratio.

\subsection{Discussions.}
\noindent \textbf{Performance considerations.} The runtime performance is an important factor in choosing a search method. The overall runtime is heavily influenced by the underlying NeRF method and the method used to produce the criteria; \textit{e.g.} for generating 250 poses on the \textit{plant\_new} scene, all methods take around 3 minutes on an NVIDIA 3090 GPU for the whole process, but EGPS takes slightly less. When it comes to the speed of the search algorithm, we see a slight lead for EGPS compared to others, averaged over all scenes; it takes around 0.8 ms to generate a pose for GRS and PIBS, whereas EGPS takes around 0.7 ms. We note that our implementations are in Python and not particularly optimized. With optimized implementations, we believe the current speed trends will still hold.

\noindent \textbf{Limitations.} We previously saw that none of the search methods worked for some scenes. Note that the reason for that might be the shortcomings of the search methods, but it can also be caused by the fact that for some scenes, training images can be the best possible images for a particular criteria. The scene exploration framework is unavoidably dependent on the accuracy of the underlying NeRF method; characteristic NeRF artefacts such as floaters and aliasing can be unavoidable at times. Some of the qualitative results show different scores (\textit{e.g.} aesthetics) for images that are quite similar; it is within the realm of possibilities for neural networks (used to generate the scores) to be influenced by changes imperceptible to human eye \cite{yucel2020deep,yucel2022robust}. Finally, the search methods sometimes fail to avoid rendering zero-density areas (\textit{e.g.} Figure \ref{fig:iqa} image on the 2nd row, penultimate column), which leads to visual artefacts.

\noindent \textbf{Future work.} We believe that the future work should first address the limitations mentioned above. Future search methods should avoid poses that would render zero-density areas (\textit{e.g.} go out of scene bounding box). Criteria functions that are much more robust to minor viewpoint changes can be investigated within the scene exploration framework for a fine-grained search performance. NeRF artefacts may be unavoidable, even with the best performing NeRF method. Therefore, we believe it would be a good direction to focus on a criteria that accurately measures NeRF artefact presence on an image, which can be used within the exploration framework to avoid such failure modes. In the same vein, extending the framework to work well with multiple criteria will be beneficial; \textit{e.g.} finding an image with Waldo that has high saliency. Finally, our scene exploration framework naturally lends itself to temporal dimension, where one could constrain the generation of a camera pose trajectory with one or multiple criteria.

\section{Conclusion} \label{sec:conclusion}
In this paper, inspired by the recent use cases of NeRF, we introduce the scene exploration framework, where the aim is to find camera poses in a scene from which one can render novel views that adhere to user-provided criteria, such as including/excluding an object, improving photo-composition and image artefact reduction. Since there are no baselines suitable for our framework, we propose GRS and PIBS baselines, and then propose EGPS which is criteria-agnostic and efficient. Testing all the methods on three tasks on various scenes shows us that EGPS performs better than the naive baselines, leading to visually better results. We hope the scene exploration framework can spark further research on NeRF-scene space exploration, and lead to new advances in content creation, multimedia production and virtual/extended reality applications.

\bibliographystyle{ACM-Reference-Format}
\bibliography{sample-base}

\end{document}